\pdfoutput=1
\documentclass[11pt]{article}

\usepackage{graphicx}
\usepackage{wrapfig}
\usepackage{subcaption}
\usepackage{float}
\usepackage[final]{acl}
\usepackage{algorithm}
\usepackage{tabularx}
\usepackage{booktabs}
\usepackage{adjustbox}
\usepackage{algpseudocode}
\usepackage{times}
\usepackage{latexsym}
\usepackage{graphicx}
\usepackage{amsmath} 
\usepackage{booktabs}
\usepackage{hyperref}
\usepackage{amssymb} 
\usepackage{multirow}
\usepackage{xcolor}
\usepackage{hyperref}       % hyperlinks
\usepackage{url}            % simple URL typesetting
\usepackage{amsfonts}       % blackboard math symbols
\usepackage{nicefrac}       % compact symbols for 1/2, etc.
\usepackage{array}
\usepackage{colortbl}
\usepackage{multirow}
\usepackage[table]{xcolor}
\usepackage{longtable}
\usepackage{todonotes}
\usepackage{xcolor}
\usepackage{todonotes}
\usepackage{makecell}
\usepackage{subcaption}
\usepackage[font=small,labelfont=bf]{caption}
\usepackage[T1]{fontenc}
\usepackage[utf8]{inputenc}

\usepackage{microtype}

\usepackage{graphicx}
\usepackage[normalem]{ulem}  % Add this in the preamble
\usepackage{listings}
\usepackage{xcolor}
\lstdefinestyle{jsonstyle}{
    basicstyle=\ttfamily\scriptsize,
    breaklines=true,
    breakatwhitespace=true,
    columns=fullflexible,
    frame=single,
    rulecolor=\color{black!40},
    xleftmargin=2pt,
    xrightmargin=2pt,
    aboveskip=2pt,
    belowskip=2pt
}
\title{From Prompt Rules to Golden Data Model: Production-Scale Generation of Personalized, Compliant Pharmaceutical Claims}

\author{
  \textbf{Suyash Mishra\textsuperscript{a}},  
  \textbf{Qiang Li\textsuperscript{b}},
  \textbf{Subhabrata Adak\textsuperscript{c}},\\
  \textbf{Upender Phogat\textsuperscript{c}},  
  \textbf{Srikanth Patil\textsuperscript{c}},  
  \textbf{Satyanarayan Pati\textsuperscript{c}},   
\\ 
  \textsuperscript{a}Roche,
  \textsuperscript{b}Accenture,  
  \textsuperscript{c}Involead,
\\
  \small{
    \textbf{Correspondence:}  
    \href{mailto:suyash.mishra@roche.com}{suyash.mishra@roche.com}},
    \href{mailto:qiang.i.li@accenture.com}{qiang.i.li@accenture.com},
    \href{mailto:subhabrata.adak@involead.com}{subhabrata.adak@involead.com}
\\ 
\small{ 
    \href{mailto:upender.phogat@involead.com}{upender.phogat@involead.com},
    \href{mailto:srikanth.patil@involead.com}{srikanth.patil@involead.com},
    \href{mailto:satyanarayan.pati@involead.com}{satyanarayan.pati@involead.com}
 }
}

\definecolor{rocheblue}{HTML}{005EB8}
\definecolor{rochelightblue}{HTML}{CCE5FF}
\definecolor{rochegray}{HTML}{F0F0F0}
\begin{document}
\maketitle

\begin{abstract}
Large language models (LLMs) are increasingly used to generate domain-specific commercial \& medical content (e.g., claims, statements, and promotional phrases) in regulated industries such as pharmaceuticals. However, ensuring its regulatory compliance, factual accuracy, and audience-specific personalization remains a major challenge. We report a production case study asking whether moving compliance and personalization signal from prompt-level rules into retrieval-time structured context improves pharmaceutical claims generation, using Composer, a deployed internal tool built on a Golden Data Model (GDM) of 1,867 pharmaceutical claims spanning four products. The GDM integrates claim metadata, references, audience labels, material type, usage history, Match Text Variant (MTV) links, and a six-dimension quality score (6C).  Across six product-audience segments, the context pipeline improves intended indication (+29\%), intended content type (+62\%) and intended target audience (+67\%), 6C quality score (+86\%), and claim reuse (+34\%). Production log analysis over the deployment period shows a 29.5\% reduction in weekly API cost. We report this as descriptive deployment evidence. This work is a systems and data-architecture contribution: it shows that in a regulated, high-stakes domain, restructuring claim data for retrieval-time filtering is a more scalable lever than expanding prompt-level rules, and documents the engineering path data cleaning, graph linking, and quality scoring Check our \href{https://storage.googleapis.com/clipsample/Composer.mp4}{Composer Video}.
\end{abstract}

\begin{figure*}[ht]
    \centering
    \vspace{-10pt}
    \begin{minipage}{1\textwidth}
        \centering
        \includegraphics[width=1\textwidth]{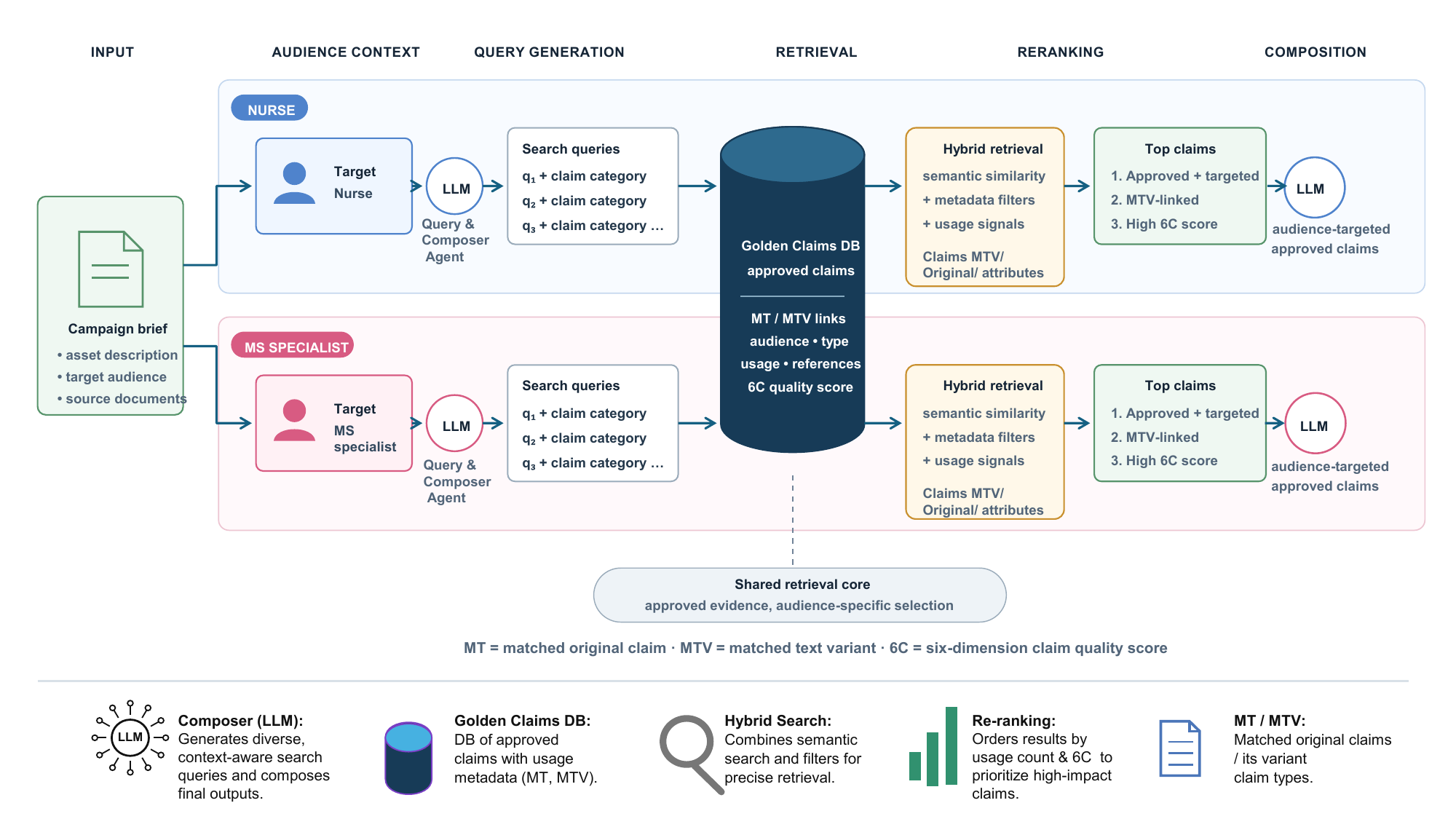} 
    \end{minipage}%
    \vspace{-10pt}
    \caption{Business Solution Map for Composer Tool Compliant Claims Generation.}
        \label{fig: architecture of LRBTC}
\end{figure*}

\section{Introduction}
AI-generated text has been detected increasingly in medical literature~\cite{wolfrath2026risingprevalencedetectedaigenerated}. A 2025 market forecast projected the AI in medical writing market to reach USD 1,760.02 million by 2030, with a CAGR of 12.52\% from 2025 to 2030~\cite{gvr2025aiMedicalWritingMarket}. However, their deployment in regulated domains such as healthcare and pharmaceuticals remains constrained by persistent concerns around hallucination, regulatory compliance, and limited controllability~\cite{hager2024evaluation}.

Pharmacy content generation is particularly challenging. Claims must be:
(i) factually grounded in approved sources,
(ii) compliant with regulatory constraints,
(iii) tailored to specific audiences such as healthcare professionals (HCPs) or patients ~\cite{barrias2024european}.

Our baseline relies on \textbf{rule-based prompting and strict templates}. In our deployment, this baseline often failed to produce personalized and contextually appropriate outputs. As observed in real-world systems, \textbf{rule-based approaches can lead to mismatches in target audience, low relevance, and overuse of inactive or general claims}.

In contrast, context-rich approaches leverage structured data, usage history, and semantic relationships to guide generation. However, the effectiveness of such approaches has not been experimentally studied. In this paper, we address following core research question:

\textbf{Does shifting compliance and personalization signals from prompt-level rules to structured retrieval-time data improve the quality and production efficiency of pharmaceutical claim generation in a deployed system?}

To answer this, we propose a golden data model schema that transforms fragmented claim data into a structured, graph-based representation with quality-aware filtering. We further evaluate rule-based and context-based generation pipelines across multiple dimensions, including compliance (e.g., prohibited words / phrases, target indication / type, MTV-Variant claims reuse), personalization (e.g. target audience), and quality (6C scores). We note that the context pipeline does not eliminate rules outright, claim-level guardrails and prohibited-word constraints still persist, but relocates them from static, product-level prompt text to claim-level, retrieval-gated metadata.

\begin{figure*}[ht]
    \centering
    \vspace{-10pt}
    \begin{minipage}{1\textwidth}
        \centering
        \includegraphics[width=0.9\textwidth]{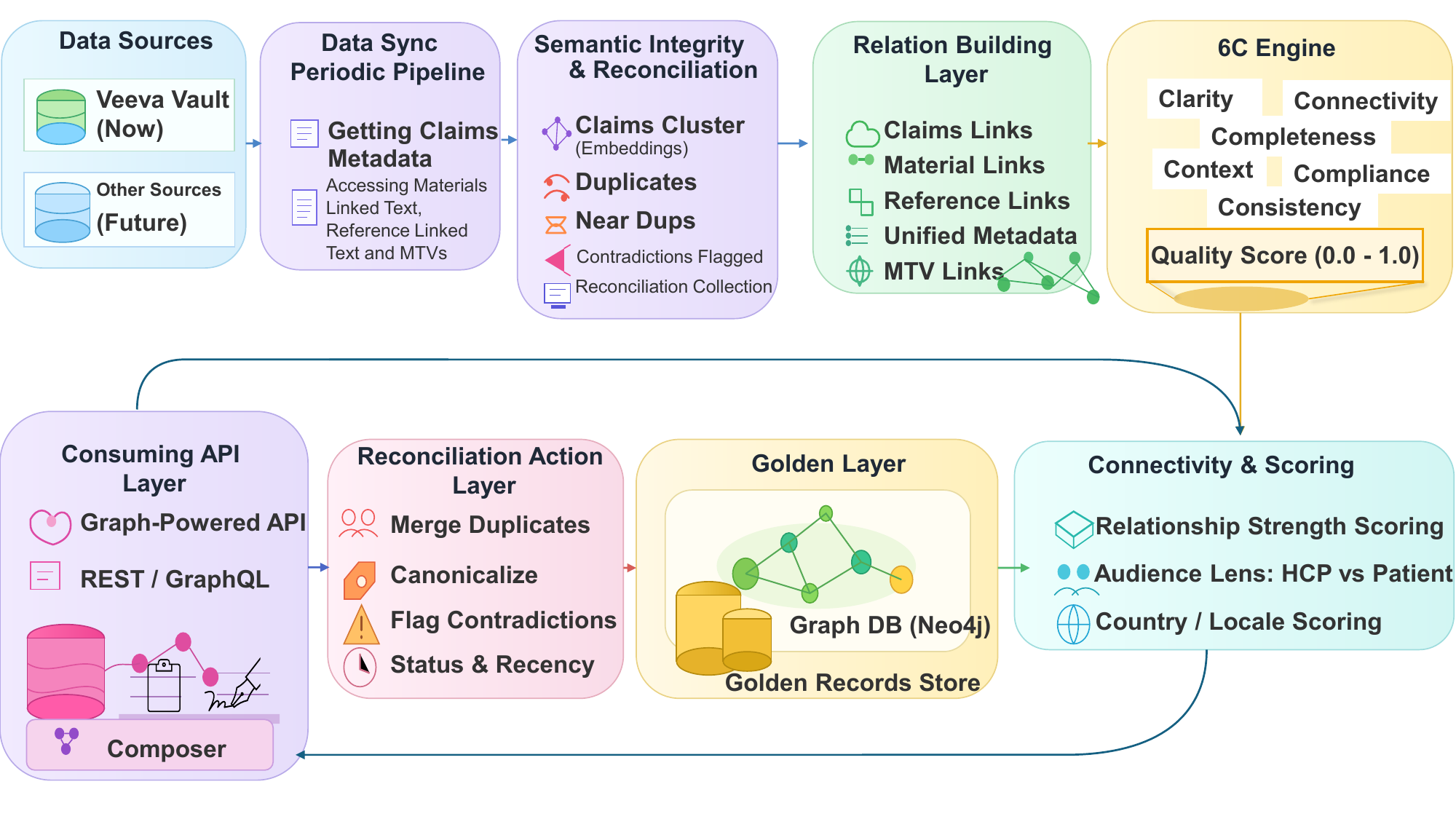} 
    \end{minipage}%
    \vspace{-15pt}
    \caption{Comprehensive Functional Layer for Compliant Claims Generation.}
        \label{fig: architecture of composer}
\end{figure*}

\section{Related Work}

%Recent work has explored controllable text generation using constraints, retrieval augmentation, and prompting strategies. 
\textbf{Pharmaceutical content management: } Veeva PromoMats supports promotional content and claims management, including substantiation and contextual claim variations~\cite{veeva2023}. IQVIA provides HCP engagement solutions~\cite{iqvia2022}. These product descriptions do not establish the internal generation architectures of the platforms. Our contribution is a production evaluation of an explicit claim-centered schema and audience-conditioned generation pipeline. Prior research has examined the informational and emotional content of prescription-drug advertising~\cite{frosch2007} and the distinction between promotion and scientific exchange~\cite{Kessler}.

\textbf{Retrieval-augmented generation (RAG) in biomedical NLP:} RAG  ~\cite{gao2023retrieval, asai2024selfrag} improves grounding by incorporating external knowledge sources.  RAG has improved factuality in knowledge-intensive generation~\cite{lewis2020}, whereas chain-of-thought prompting has improved arithmetic, commonsense, and symbolic reasoning~\cite{wei2022}; the latter does not by itself establish reduced hallucination. However, standard RAG systems lack domain-specific constraints such as regulatory compliance and structured claim validation. RAG alone is insufficient for pharmaceutical claim generation because it does not enforce the audience-specific stylistic transformation that distinguishes a KOL-facing mechanism statement from a patient-facing aspirational statement derived from the same input.
Prior work in biomedical NLP, has mostly focused on information extraction and summarization, but less attention has been given to controlled, personalized generation under strict regulatory requirements. Our work differs in two key aspects:
(1) we introduce a data-centric framework for controlling generation,
and (2) we provide an empirical comparison between rule-based and context-based approaches in real-world setting.

\textbf{Verification and hallucination evaluation:}  
Recent benchmarks such as \textit{SciVer}~\citep{wang2025sciver} evaluate multimodal scientific claim verification with expert-annotated supporting evidence. HonestVQA~\citep{anonymous2025the} studies confidence calibration in document visual question answering, while Decomposed NLI (DNLI)~\citep{Yanuka_2025} evaluates individual propositions in image captions. MLLMGuard~\citep{gu2024mllmguard} assesses multimodal safety across privacy, bias, toxicity, truthfulness, and legality. Industrial traffic-incident benchmarks also examine LLM hallucination~\citep{li-etal-2025-llms}. These studies concern distinct tasks and do not directly validate pharmaceutical claim generation.

\section{Methodology}

Given a target audience (e.g., HCP, Patient), content type (e.g., Email, Banner), and a set of claims, the goal is to generate compliant and personalized text. The rule-based approach relies on predefined instructions, including: strict citation rules, verbatim claim usage constraints, and template-based generation. While this ensures high compliance, it limits flexibility and personalization. We propose a golden data model schema that enhances claim usability via 4 key components as shown in Fig \ref{fig: architecture of composer}:

\textbf{1. Semantic Integrity Layer:}  periodic pipeline executing high-volume "upserts" from Veeva Vault and other sources into the central library.
Claims are clustered using keywords and embeddings to identify duplicates, near-duplicates, contradictions. 

\textbf{2. Graph-Based Representation:} Claims are linked to: references, materials, audience metadata,
forming a relational graph structure (as shown in Appendix Figure \ref{fig: architecture of graph}. This is the relationship building layer. Establishes the structural links between Claims, References, and Materials. It aggregates metadata to provide a unified view of derivation and utilization, stored in Graph DB.

\textbf{3. 6C Quality Framework:} Each claim is then scored along six dimensions:
\textit{Clarity, Completeness, Consistency, Connectivity, Compliance, Context}, as shown in Table \ref{tab:6c_unified_rubric}  \textbf{how} weights differ, Table \ref{tab:veeva_failure_modes} \textbf{why} each dimension is scored this way, Table \ref{fig:what} \textbf{what} it evaluates. It translates relational evidence into a mathematical "Strength Score", quantifies how firmly a claim to its supporting scientific text.

\textbf{4. Alignment Annotation Protocol}. Indication, content-type, and target audience alignment were then judged by internal US medical-content writers team not involved in system development, using binary match against the campaign brief's stated indication/content-type/ audience fields. Judgment was BLINDED  to pipeline (Pre vs. Post).

\textbf{5. Audience Lens:} Claims are filtered and ranked based on target audience, region, and usage patterns. This is the reconciliation action layer - the final decision gate. Based on 6C scores, status, and recency, it executes the "Golden Actions" (Merge/Suppress/Flag), publishing the final, high-integrity records to the Gold Layer Graph DB.

In our context-based approach, the LLM then retrieves top-ranked claims, applies semantic filtering and ranking, generates outputs conditioned on high-quality, audience-specific claims.

\begin{figure}[ht]
    \centering
    \includegraphics[width=1.05\columnwidth]{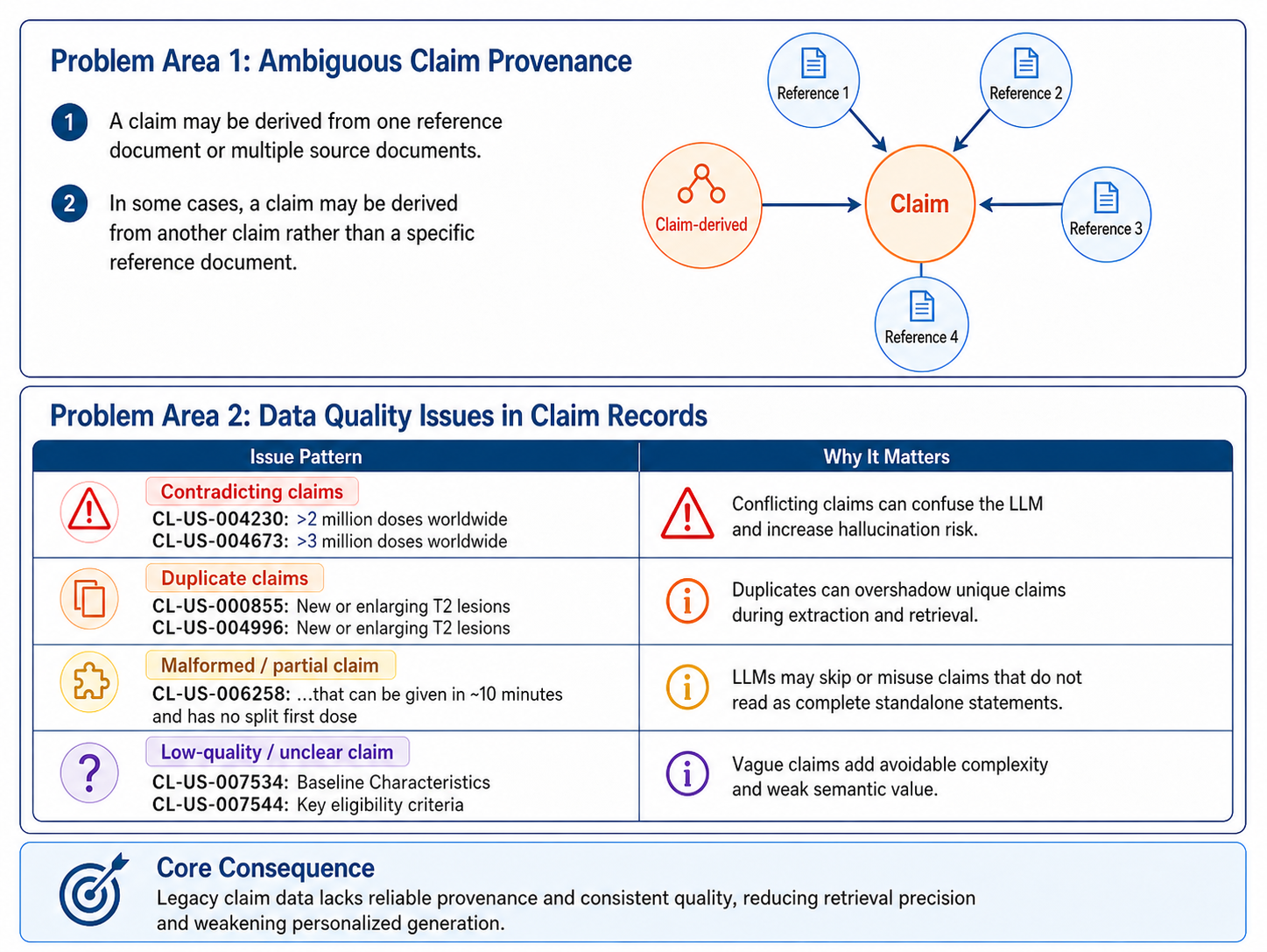}
    \vspace{-6pt}
    \caption{Failure modes in legacy Veeva claim data. We observe two primary problem areas: ambiguous provenance, where the source of a claim is not always uniquely grounded, and heterogeneous data quality issues, including contradictory, duplicate, malformed, and low-value claims. These failure modes introduce retrieval ambiguity and reduce the effectiveness of downstream claim personalization.}
      \label{fig:segment_improvement}
    \vspace{-8pt}  
\end{figure}

\begin{table*}[ht]
\centering
\vspace{-20pt}
\caption{Golden Data Model Schema for claim-level records. The schema integrates cleaned raw claim data with LLM-derived tagging layers and 6C quality scores. It also extracts tone signals and constraint-aware key phrases, enabling more controlled, compliant, and personalized downstream claim generation.}
\small
\begin{tabular}{p{3.5cm}p{2cm}p{9cm}}
\toprule
\textbf{Field} & \textbf{Type} & \textbf{Description} \\
\midrule
Claim ID & String & Unique claim identifier, e.g., \textbf{\texttt{CL-US-000754}}. \\
Create Date & Datetime & Timestamp when the claim record was created. \\
Modified Date & Datetime & Timestamp when the claim record was last modified. \\
Product & String & Product associated with the claim, e.g., \textbf{Ocrevus}. \\
Country & String & Country or market where the claim is approved. \\
\textbf{Match Text } & Text & Original approved claim text. e.g., \textbf{OCREVUS is proven effective in reducing relapses in RMS, slowing disability progression in RMS.}\\
Claim Category & Categorical & Business or medical category of the claim, e.g., \textbf{Disease Education, Efficacy, Safety, Dosing.} \\
Scientific Area & String & Scientific or therapeutic area, e.g., \textbf{Neuroscience}. \\
Indication & String & Disease or indication linked to the claim, e.g., \textbf{Multiple Sclerosis}. \\
Lifecycle State & Categorical & Approval or lifecycle status, e.g., \textbf{Approved, Archived, Draft.} \\
Source Claim ID & String & Optional parent or source claim identifier. \\
Reference IDs & List[String] & Linked reference identifiers supporting the claim. \\
Reference Texts & List[Text] & Extracted reference snippets linked to the claim. \\
Content Type & Categorical & Material or asset type, e.g., \textbf{Slide Deck, Email, Website.} \\
\textbf{Target Audience} & Categorical & Intended audience, e.g., \textbf{Patient, HCP, Nurse}. \\
\textbf{AI Guardrails Hallucination Prevention} & JSON/List & \textbf{Claim-specific rules derived from match text, reference text, audience, material type, and claim category}. Used to prevent unsupported modification or hallucination. \\
\textbf{Claim Specific Business Rule} & JSON & Marketing vector store metadata, including\textbf{ tone signals and primary message type.} \\
\textbf{Prohibited Words Negative Constraints} & JSON & \textbf{Important anchor phrases, prohibited transformations, and phrases to avoid or not overstate.} \\
\textbf{6C Score} & Float & Quality score from 0.0 to 1.0 based on \textbf{Clarity, Completeness, Consistency, Connectivity, Compliance, and Context.} \\
\bottomrule
\end{tabular}

\label{tab:golden_data_schema}
\end{table*}

\section{Dataset and Experimental Settings}
We construct our dataset of more than \textbf{1,800} pharmaceutical claims across \textbf{4} products: Hemlibra, Ocrevus, Ocrevus Zunovo, and Vabysmo. The claims span \textbf{11} categories, including disease education, efficacy, safety, dosing and administration, clinical practice guidelines, and patient access. All claims are retrieved from Veeva and are associated with metadata such as references, usage history, and audience annotations. We evaluate the two pipelines using Claude Sonnet 4.5, the production model for Composer; we additionally report per-token cost benchmarks for Gemini 2.5 Pro/Flash, the deployment comparison in §5 uses Sonnet 4.5 throughout ~\cite{google2026gemini25pro, anthropic} across multi-dimensions: \textbf{1. Compliance Rate \& Claim Utilization}: The percentage of outputs that correctly use approved claims, the proportion of relevant claims used, and content-type alignment with intended format. \textbf{2. Personalization Accuracy \& 6C Score}: Alignment with target audience and the average quality of selected claims measured by 6C scoring (\ref{sec:A2}). \textbf{3. Deployment API Cost \& Latency}: Request-level scalability metrics, including success rate, average latency, P50 latency, request tokens per week, total cost, and tokens per request.

\section{Main Results}

\begin{table*}[t]
\centering
\scriptsize
\setlength{\tabcolsep}{3.5pt}
\caption{Overview of claims, references, materials, and MTV (Match Text Variation, i.e. variant claims) availability and utilization across products. A total of \textbf{1,867} claims are identified across (Hemlibra, Ocrevus, Ocrevus Zunovo, Vabysmo) products for the US market.
72\% of claims are utilized in materials, indicating strong adoption in content. But, for Ocrevus Zunovo it is only 37\%.
98\% of claims are Sourced from references, while the remaining 2\% do not have any linked reference.
   }
\small
\begin{tabular}{lcccccccccc}
\toprule
\textbf{Product} 
& \textbf{\#Claims} 
& \textbf{\#Refs} 
& \textbf{\#Materials} 
& \textbf{\%Used} 
& \textbf{\%With Ref} 
& \textbf{\#MTV} 
& \textbf{\#MTV Used} 
& \textbf{\%MTV} 
& \textbf{\%MTV Used} \\
\midrule
Hemlibra & 745 & 721 & 593 & 80\% & 97\% & 297 & \textcolor{red}{\textbf{187}} & 39.87\% & 25.10\% \\
Ocrevus & 357 & 355 & 215 & 60\% & 99\% & 113 & 31 & \textcolor{red}{\textbf{31.65}}\% & \textcolor{red}{\textbf{8.68}}\% \\
Ocrevus Zunovo & 122 & 122 & 45 & 37\% & 100\% & 29 & 8 & 23.77\% & 6.56\% \\
Vabysmo & 643 & 640 & 489 & 76\% & 100\% & 358 & 102 & \textcolor{red}{\textbf{55.68}}\% & 15.86\% \\
\midrule
\textbf{Total} & 1867 & 1838 & 1342 & 72\% & 98\% & 797 & 328 & \textcolor{red}{\textbf{42.69}}\% & \textcolor{red}{\textbf{17.57}}\% \\
\bottomrule
\label{fig:claims}
\end{tabular}

\end{table*}

Our results show that context-based generation outperforms rule-based pipeline, Fig.  \ref{fig:legacy_rule_json} Pre-pipeline legacy rule diagram.

\textbf{Compliance \& Personalization: }Context-based generation achieves near-perfect compliance across all segments, compared to lower compliance in rule-based setups. In the existing Pre-setup "rule-based pipeline" only 71\% of the claims are aligned with the \textbf{intended indication}, only 38\% of the claims are aligned with the \textbf{intended content type}, only 33\% of the claims are aligned with the \textbf{intended target audience}. However, as shown \textbf{Table \ref{fig:preandpost}}, post context pipeline 100\% claims are matching with intended indication, intended content type and intended target audience. Furthermore, examining \textbf{Table \ref{fig:product}} shows that this improvement generalizes across \textbf{product level}, specifically for Vabysmo and Ocrevus, as well as across \textbf{role types}, including HCPs, nurses, and patients’ physicians.

We distinguish two classes of metric in this evaluation. Structurally guaranteed metrics — indication, content-type, and target-audience alignment, and "approved claims used" — are enforced by the Post pipeline's retrieval filters by construction: Composer only retrieves claims that already match the requested indication/content-type/audience, so 100\% alignment on these dimensions reflects the filter design. We report them because the Pre pipeline's failure to enforce the same constraints (33–71\%). Learned or judged metrics — MTV utilization, 6C quality score of selected claims are not guaranteed by construction and are the metrics that actually test whether structured context improves outcomes.

Alignment with target audience rate improves substantially, with context-based pipeline, the content generation (Post) achieved 100\% compliance with \textbf{approved claims}, compared to \textbf{71\% }in previously created content.

\begin{table}[ht]
\centering
\vspace{-10pt}
\caption{Performance comparison between rule-based (Pre) and context-based (Post) systems. Here, we reports MTV utilization within the Pre/Post evaluation sample specifically.}
\small
\begin{tabular}{p{3.5cm}ccc}
\toprule
\textbf{Metric} & \textbf{Pre} & \textbf{Post} & $\boldsymbol{\Delta}$ \\
\midrule
Approved Claims  & 71\% & 100\% & +29\% \\
MTV Utilization & 0\% & 34\% & +34\% \\
Indication Alignment & 71\% & 100\% & +29\% \\
Content-Type Alignment & 38\% & 100\% & +62\% \\
Target Audience Alignment & 33\% & 100\% & +67\% \\
\bottomrule

\label{fig:preandpost}
\end{tabular}

\end{table}

\textbf{Claim Utilization:} as shown in Table \ref{fig:claims} and Table  \ref{tab:example_claim_full}, the use of high-quality claims increases, particularly for previously \textbf{underutilized claim variants}. Raw claim repositories are structurally rich but operationally underutilized without contextual modeling, e.g., Ocrevus: 31.65\% MTV (Variants of Claim) exists but only 8.68\% used. High claims coverage but low MTV usage with avg only 17.68\%. Vabysmo leads in MTV coverage (55.68\%) with relatively higher linkage, Hemlibra has the highest absolute MTV usage (187 used in materials), Ocrevus \& Ocrevus Zunovo show low utilization, indicating a higher reliance on direct claim text instead of MTVs (Only 6.56 \% - 8.68\%). Overall, MTV availability is 42.69\%, while only 17.68\% are utilized in materials, highlighting a significant usage gap. Further shown in Table \ref{fig:preandpost}, after the context pipeline, the 
\textbf{MTV claim utilization} increased 34\%, unlocking previously unused claim  across all segments (see Table \ref{fig:product}: MTV).

\begin{table}[ht]
\centering
\vspace{-5pt}
\caption{Comparison of API cost, latency, and token-usage. The context pipeline reduces weekly API cost by 29.5\%, lowers total weekly token usage, and substantially decreases response-token generation by 65.4\%. It also improves request success rate from 99.2\% to 100\%. Although the average cost per request increases from 0.0234 dollar to 0.0316 due to richer contextual retrieval, the overall weekly cost decreases because it improves first-pass usefulness, decreases response-token generation, and reduces the number of repeated requests.}
\scriptsize
\setlength{\tabcolsep}{1pt}
\renewcommand{\arraystretch}{1}
\begin{adjustbox}{width=0.5\textwidth}
\begin{tabular}{lccc}
\toprule
\textbf{Metric} & \textbf{Pre: Rule-heavy Pipeline} & \textbf{Post: Context Pipeline} & \textbf{Change} \\
\midrule
\multicolumn{4}{l}{\textit{Model configuration}} \\
Model provider & Anthropic & Anthropic & -- \\
Model & Claude Sonnet 4.5 & Claude Sonnet 4.5 & -- \\

\midrule
\multicolumn{4}{l}{\textit{Request-level scalability}} \\
\textbf{Total requests per week} & \textbf{5,485} & \textbf{2,871} & $\downarrow$ 47.7\% \\
\textbf{Success rate} & 99.2\% & 100\% & $\uparrow$ 0.8 \%  \\
Average latency & $\sim$2.8k ms & $\sim$3.0k ms & $\uparrow$ 7.1\% \\
Latency (P50) & 2,100 ms & 1,956 ms & $\downarrow$ 6.9\% \\
Maximum latency & 2,909.07 ms & 3,819.86 ms & $\uparrow$ 31.3\% \\

\midrule
\multicolumn{4}{l}{\textit{Token and cost efficiency}} \\
Total tokens / week & $\sim$31.4M & $\sim$26.2M & $\downarrow$ 16.6\% \\
\textbf{Request tokens / week} & $\sim$28.42M & $\sim$25.2M & $\downarrow$ 11.3\% \\
\textbf{Response tokens / week} & $\sim$2.92M & $\sim$1.01M & $\downarrow$ \textbf{65.4}\% \\
\textcolor{red}{\textbf{Total cost / week }}& \$128.70 & \$90.69 & $\downarrow$ \textcolor{red}{\textbf{29.5}\%} \\
Tokens / request & $\sim$5,714 & 9,126 & $\uparrow$ 59.7\% \\
\textbf{Average cost / request }& $\sim$\$\textbf{0.0234} & $\sim$\$\textbf{0.0316} & $\uparrow$ 35.0\% \\

\bottomrule
\end{tabular}
\end{adjustbox}
\label{tab:pre_post_cost_latency}
\end{table}

\begin{table}[t]
\centering
\vspace{-10pt}
\caption{Quality vs. usage comparison before/after the Golden Data Model. C  number of claims, U number of re-used claims}
\footnotesize
\setlength{\tabcolsep}{2.5pt}
\begin{tabular}{p{2.8cm}lccccc}
\toprule
\textbf{Segment} & \textbf{Stage} & \textbf{C} & \textbf{U} & \textcolor{red}{\textbf{Avg 6C score}} & \textbf{Min} & \textbf{Max} \\
\midrule
OCR-MS-Nurse & Pre  & 5 & 2 & 0.83 & 0.83 & 0.83 \\
              & Post & 5 & 5 & 0.91 & 0.83 & 1.00 \\

OCR-MS-Patient & Pre  & 3 & 3 & 0.91 & 0.83 & 1.00 \\
               & Post & 6 & 6 & 0.93 & 0.83 & 1.00 \\

\textcolor{red}{\textbf{VAB-DME-Patient}} & Pre  & 2 & \textbf{0} & 0.00 & 0.00 & 0.00 \\
                & Post & 3 & \textbf{3} & 0.86 & 0.83 & 0.89 \\

VAB-RVO-HCP & Pre  & 5 & 4 & 0.87 & 0.83 & 0.90 \\
            & Post & 5 & 5 & 0.87 & 0.79 & 0.90 \\

VAB-nAMD-Physician & Pre  & 5 & 4 & 0.88 & 0.85 & 0.89 \\
                   & Post & 6 & 6 & 0.91 & 0.88 & 0.95 \\
\bottomrule
\end{tabular}
\label{fig:Quality vs. usage comparison}
\end{table}

\paragraph{Quality \& 6C:}
As shown in Table \ref{fig:Quality vs. usage comparison}, the Average 6C scores are higher in the context-based approach, indicating improved clarity, consistency, and contextual relevance. POST dominates PRE in almost all segments, especially: VAB DMe Patient: 0.0 \ensuremath{\rightarrow} 0.86, OCR MS Nurse: 0.83 \ensuremath{\rightarrow} 0.91, VAB-DME-Patient: 0 re-used claims to all re-used claims. This is clear evidence our golden data model improves quality.
With the higher-scoring 6C claims are used in content generation, we also see the cost reduced, as shown in left side Table \ref{tab:pre_post_cost_latency},
the \textbf{deployed context pipeline substantially reduces response-token generation by 65.4\%, suggesting cleaner and more controlled outputs}. It also improves the request success rate from 99.2\% to 100\%. In practice, the total cost is not determined only by the cost of a single generation call, but also by user behavior and re-generation frequency. \textbf{When generated 
outputs are better aligned with the user's intent, target audience, and content constraints, users may need fewer repeated requests to obtain satisfactory results.} Therefore, the cost profile is not a simple binary outcome. If the context pipeline were used at the same request volume as Pre, the total cost would increase by roughly 35\%, consistent with the higher average cost per request. In the observed deployment, however, the number of requests decreased from 5,485 to 2,871, resulting in a lower total weekly cost.

\begin{table}[t]
\centering
\vspace{-10pt}
\caption{\textbf{Segment-level} performance comparison. Metrics: Used = reused claims, MTV = variation usage, Ind = indication, Type = content-type, Aud = audience.}
\scriptsize
\setlength{\tabcolsep}{2pt}
\begin{tabular}{lcccccccccl}
\toprule
\textbf{Input}
& \multicolumn{2}{c}{\textbf{Used}}
& \multicolumn{2}{c}{\textcolor{red}{\textbf{{MTV}}}}
& \multicolumn{2}{c}{\textbf{Ind}}
& \multicolumn{2}{c}{\textbf{Type}}
& \multicolumn{2}{c}{\textbf{Aud}} \\
\cmidrule(lr){2-3}\cmidrule(lr){4-5}\cmidrule(lr){6-7}\cmidrule(lr){8-9}\cmidrule(ll){10-11}
& Pre & Post & Pre & Post & Pre & Post & Pre & Post & Pre & Post \\
\midrule
VAB-RVO-HCP        & 80 & 100 & 0 & 20 & 20 & 100 & 40 & 100 & 60 & 100 \\
VAB-nAMD-Physician & 80 & 100 & 0 & 17 & 60 & 100 & 20 & 100 & 40 & 100 \\
VAB-ALL-Patient    & 100 & 100 & 0 & 43 & 100 & 100 & 100 & 100 & 50 & 100 \\
VAB-DME-Patient    & 0 & 100 & 0 & 33 & 50 & 100 & 0 & 100 & 0 & 100 \\
OCR-MS-Nurse       & 40 & 100 & 0 & 40 & 100 & 100 & 20 & 100 & 0 & 100 \\
OCR-MS-Patient     & 100 & 100 & 0 & 50 & 100 & 100 & 33 & 100 & 33 & 100 \\
\midrule
\textbf{Total avg. in \%}     & 71 & 100 & 0 & 34 & 71 & 100 & 38 & 100 & 33 & 100 \\
\bottomrule
\end{tabular}

\label{fig:product}
\end{table}

\begin{figure}[ht]
    \centering
    \includegraphics[width=\columnwidth]{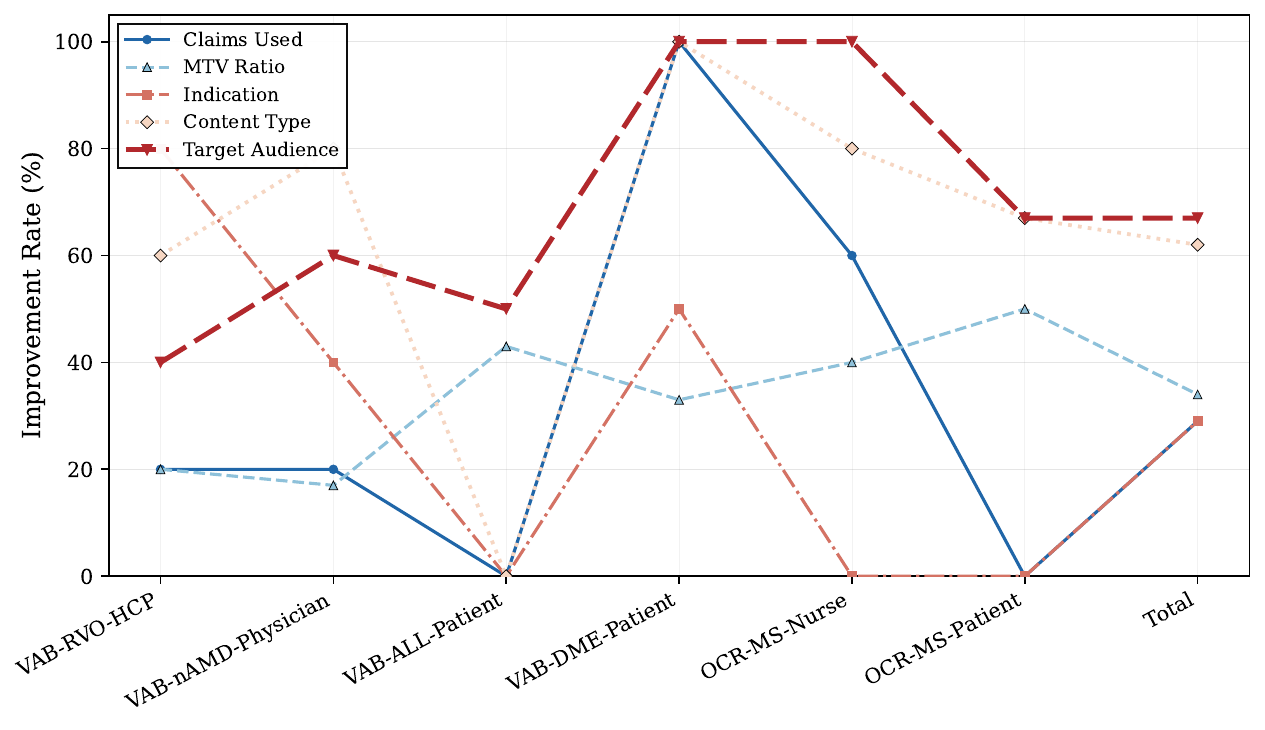}
    \vspace{-20pt}
    \caption{Improvement rates from the rule-based baseline to the Golden Data Model across customer segments. We report gains in claim usage, MTV utilization, indication alignment, content-type alignment, and target-audience alignment. The most pronounced improvements occur in \textbf{audience and content alignment}, suggesting that structured contextual retrieval better supports personalized claim generation.}
      \label{fig:segment_improvement_appendix}
     
\end{figure}

In addition, with Figure \ref{fig:segment_improvement2} and Appendix  Table \ref{tab:veeva_failure_modes}, it is suggested that rule-based constraints alone are insufficient for complex generation tasks. While they enforce correctness, they fail to capture contextual nuances required for personalization. In contrast, structured contextual data enables the model to select more relevant claims, adapt to audience-specific needs, and maintain compliance via data filtering rather than prompt constraints. This also highlights the importance of data-centric approaches in controlling LLM behavior.

\begin{table}[h!]
\centering

\caption{Issue Examples in Veeva Vault claims database.} 
\setlength{\tabcolsep}{1.5pt}
\renewcommand{\arraystretch}{1.5}
\begin{adjustbox}{width=0.5\textwidth}
\begin{tabular}{p{2cm} p{5cm} p{3cm}}
\toprule
\textbf{Issue} & \textbf{Example} & \textbf{Impact} \\
\midrule
Contradicting Claims 
& (CL-US-004230) $>$2M doses vs (CL-US-004673) $>$3M doses 
& Confusion for LLM, increases hallucination risk \\
\midrule
Duplicate Claims 
& (CL-US-000855) vs (CL-US-004996) same T2 lesion claim 
& Over-representation, suppresses diversity \\
\midrule
Malformed Claims 
& (CL-US-006258) starts with incomplete sentence 
& Violates brand rules, reduces usability \\
\midrule
Low-quality Claims 
& (CL-US-007534) “Baseline Characteristics” 
& Adds noise, no semantic value \\
\bottomrule
\end{tabular}
\end{adjustbox}
\end{table}

\section{Conclusion}

This work demonstrates that structured context moving compliance and personalization signal from prompt-level rules into retrieval-time claim metadata provides a more scalable and effective architecture for content generation in a regulated industry than expanding prompt-level rules alone. We evaluated across \textbf{compliance, target-audience, indication, content-type alignment, MTV reuse, latency, token usage, request success rate, and API cost}. We also formalized 6C Algorithm for evaluating claim quality across Clarity, Completeness, Consistency, Connectivity, Compliance, and Context Availability. Our results show that the context-based pipeline improves compliance while reducing total weekly cost. Although richer retrieval increases the average cost per request, the overall system becomes more efficient - cleaner outputs, and a higher success rate. This work demonstrates that structured context rather than explicit rule instructions provides a more effective LLM solution for regulated industry.

%We introduced further a data-centric, graph-grounded claim representation can replace brittle prompt compliance control by moving compliance and personalization signals into retrieval-time structure.

%\textbf{Limitation:}
%Despite being the first to release such large industrial dataset on accident information, our data still have limitations. To more effectively test the temporal and spatial awareness capabilities of LLMs, we need to manually annotate more spatial and temporal data and ground truths. Expanding to other regions or cities would require additional approvals from governments or institutions, which could further enhance our dataset.
%\textbf{Future work:}
%To address these limitations, we will continually collect accident information from various cities. Additionally, we plan to exploring various other functionalities of LLMs beyond just hallucinations.

%\section{Acknowledgments}  
%We sincerely thank Samik Adhikary and Puneet Srivastava for their sponsorship support from Roche. We also appreciate the insightful discussions and technical assistance provided by Janina Kummerfeldt, Philipp Müller, Kathrin Schwan from Accenture, as well as Aditya Kumar from Involead.  

\section{Limitations and Ethic Statements}

This study focuses on a real-world pharmaceutical content generation setting, where regulatory compliance, claim traceability, and audience-specific adaptation are central requirements. The dataset reflects approved claim assets from selected pharmaceutical products and specific market context, which enables detailed analysis of compliance-sensitive generation. Due to the regulated and proprietary nature of the underlying claim data, we cannot release the full production dataset. To support transparency and reproducibility, we also provide representative cleaned examples, schema-level records, scoring rubrics, and system-level descriptions, logs, tool videos in the appendix. The observed efficiency gains are based on production log analysis. These results should therefore be interpreted as deployment-level evidence rather than a purely model-intrinsic efficiency claim. 

Our current experiments focus on pharmaceutical compliance because it represents a high-stakes domain where controllable and evidence-grounded generation is especially important. Future work will extend the proposed data-centric approach to additional regulated domains, such as financial services and manufacturing, to further assess its generalizability across different compliance and content-governance settings.

Composer generates promotional and medical content for real pharmaceutical products.All generated content passes through existing MLR (Medical, Legal, Regulatory) review before reaching end users; Composer does not publish directly to patients or HCPs. LLM-derived guardrails and prohibited-phrase constraints (Table 1) are validated and derived from and cross-checked against the original approved claim text at generation time, with human review flagged in the schema. Provenance and approval status are preserved through canonicalization process when records are reconciled in the Golden Layer.

\section{Acknowledgments}  

We sincerely thank Samik Adhikary and Puneet Srivastava for their sponsorship support from Roche. We also appreciate the insightful discussions and technical assistance provided by Qiang Li from Accenture; Jennifer McGuire, Christopher Chu, Alex Connor, and Liz Stutz for business support from Roche; and Upender Phogat and Akshat Gupta for US MLP Composer development support from Involead. ~\cite{li-etal-2025-llms, mishra2026llmsbetteronestudentteacher}

This Composer use case, as well as the GenAI platform \citep{mishra-etal-2026-understanding} \citep{mishra-etal-2026-scaling}, would not have been possible without their contributions. We further extend our gratitude to the backend engineering teams who supported development, and to the healthcare professionals (HCPs), testers, and Roche Lab users whose consistent feedback brought our use cases, particularly the Agent to life and enabled continuous improvement.

This paper represents our academic contribution, in which we formalize experiments and evaluation methodologies using academic benchmarks. Through this work, we aim to share industry lessons learned and report valuable large-scale GenAI experiments in the pharmaceutical domain.

\appendix
\clearpage
\newpage
\section{Appendix}
\label{sec:appendix}

In this section we provide the supplementary compiled together with the main paper includes:
\begin{itemize}

\item Ablation study of the framework’s performance in terms of token usage, latency, and monetary cost, mathematical rationale of 6C, the 6C Claim Quality Scoring algorithm, and associated loss functions;

\item Property dataset including the example claim record (CL-US-002291) represented in the Golden Data Model Schema format, and a comprehensive claims graph view;
\item Failure mode matrix for Veeva claim records. The table summarizes observed data quality and governance pain points that motivate the Golden Data Model;
\item Legacy product-level rule JSON used in the rule-based Composer pipeline;

\item The prompt instruction lists;

\end{itemize}

\subsection{Ablation study: framework perform in terms of token count, latency, and monetary cost }

In addition, we record per-request token usage and billing of production logs from different model providers. Gemini 2.5 Flash requests cost sub-cent (e.g., ~0.38–0.64¢ for ~1.8–1.9K tokens), while Gemini 2.5 Pro typically costs a few cents per request. For Anthropic endpoints, observed costs were also low (e.g., ~1.46–3.93¢ for ~4–11.9K tokens; ~0.32–0.37¢ per 1K tokens), which enables scalability for industry application.

The model used in Table ~\ref{tab:pre_post_cost_latency} both settings is \texttt{us.anthropic.claude-sonnet-4-5}. 
The cost comparison reveals an interesting trade-off. After integrating the context pipeline, The observed weekly API cost fell from $128.70 to $90.69 (\ensuremath{-}29.5\%) across the measurement window, driven by a drop in request volume (5,485 \ensuremath{\rightarrow} 2,871) that outweighs the 35\% higher average cost per request. We hypothesize this reflects fewer \textbf{re-generation} requests as output quality improves, consistent with the 65.4\% drop in response-token generation. Here we did not measure regeneration rate, active-user counts, or campaign mix directly. We therefore report this as descriptive deployment evidence of an association between the context pipeline and lower weekly spend.

However, the \textbf{deployed context pipeline substantially reduces response-token generation by 
65.4\%}. It also improves the request success 
rate from 99.2\% to 100\%. In practice, the total cost is not determined only by the cost of a 
single generation call, but also by user behavior and re-generation frequency. \textbf{When generated 
outputs are better aligned with the user's intent, target audience, and content constraints, users 
may need fewer repeated requests to obtain satisfactory results.}

Therefore, the cost profile is not a simple binary outcome. This suggests that \textbf{business-level efficiency depends jointly on API cost per request, request 
volume, success rate, and perceived output quality}.

\begin{table}[t]
\centering
\caption{Example per-request costs from system logs. Costs depend on provider/model tier, motivating cost-aware routing in multi-pass verification.}
\small
\begin{tabular}{l l  }
\hline
Provider/Model  & Cost / 1K tokens \\

\hline
Gemini 2.5 Pro   & 1.41\textcent  -- 1.49\textcent \\
Gemini 2.5 Flash & 0.21\textcent  -- 0.34\textcent \\
Anthropic (Claude) & 0.32\textcent -- 0.37\textcent \\

\hline
\end{tabular}

\label{tab:cost1}
\end{table}

\subsection{Ablation study: the 6C Scoring Framework }
\label{sec:A2}
The 6C Scoring Framework evaluates claims across six dimensions (the "6 C's"). Each dimension scores between 0.0 and 1.0. Connectivity is evaluated first based on the presence of references, materials, and archived references. Compliance is evaluated next, it requires that connectivity has passed (score >= 0.9) and checks life-cycle approval status. Context Availability is evaluated only after both connectivity and compliance have been established. Clarity, Completeness, and Consistency are evaluated in parallel via LLM calls.

The scoring pipeline follows these steps:
\begin{itemize}
\item  Data retrieval: Claims and MTV (Match Text Variant) data are fetched from the Neo4j graph database ~\cite{robinson2015graph} along with their linked References and Materials.

\item  Connectivity scoring: Each claim is scored based on the presence of its linked references, archived references, and materials.

\item  Compliance scoring: Claims that pass connectivity are checked for life-cycle approval status.

\item  Parallel LLM evaluation: Three concurrent API calls to Claude Sonnet 4 evaluate Clarity, Completeness, and Consistency. LLM-based assessment is motivated by prior evaluation studies~\cite{zheng2023judging, liu2023geval}; Claude Sonnet 4 is documented separately~\cite{anthropic4}.

\item  Context Availability scoring: Computed deterministically from metadata column availability (requires connectivity and compliance first).

\item  Final aggregation: The 6C Score is the weighted arithmetic mean of all six dimension scores.

\end{itemize}

Here, for LLM Configuration Model,we use us.anthropic.claude-sonnet-4-20250514-v1:0 (AWS Bedrock) ~\cite{anthropic4}, Temperature we set 0.0 (low-temperature sampling)and Max Tokens: 65,000, Concurrency: 40 parallel evaluations.

For each claim $x_i$, we compute a six-dimensional quality vector:
\[
\mathbf{s}_i =
[
s_i^{\mathrm{clar}},
s_i^{\mathrm{comp}},
s_i^{\mathrm{cons}},
s_i^{\mathrm{conn}},
s_i^{\mathrm{compl}},
s_i^{\mathrm{ctx}}
],
\]
where each dimension score is bounded in $[0,1]$. The final 6C score is a weighted arithmetic mean:
\[
S_{6C}(x_i) = 
\sum_{k=1}^{6} w_k s_i^{(k)},
\quad
\sum_{k=1}^{6} w_k = 1,
\quad
s_i^{(k)} \in [0,1].
\]
In our experiments, we assign equal weights to all six dimensions:
\[
w_k = \frac{1}{6}.
\]
Thus:
\[
\begin{aligned}
S_{6C}(x_i) =
\frac{1}{6}\Big(
& s_i^{\mathrm{clar}} +
s_i^{\mathrm{comp}} +
s_i^{\mathrm{cons}} \\
& +
s_i^{\mathrm{conn}} +
s_i^{\mathrm{compl}} +
s_i^{\mathrm{ctx}}
\Big).
\end{aligned}
\]

\begin{table}[t]
\centering
\small
\setlength{\tabcolsep}{4pt}
\begin{tabular}{p{2.1cm}p{3.1cm}p{1.5cm}}
\toprule
\textbf{Dimension} & \textbf{Evaluation Method} & \textbf{Range} \\
\midrule
Clarity & LLM, 7 weighted heuristics & $[0,1]$ \\
Completeness & LLM, 5 weighted heuristics & $[0,1]$ \\
Consistency & LLM, 4 weighted heuristics & $[0,1]$ \\
Connectivity & Data-driven linkage checks & $[0,1]$ \\
Compliance & Lifecycle and connectivity checks & $[0,1]$ \\
Context Availability & Metadata availability checks & $[0,1]$ \\
\bottomrule
\end{tabular}
\caption{Overview of the 6C scoring dimensions. Each dimension is normalized to $[0,1]$ and combined into the final 6C score.}
\label{tab:6c_overview}
\end{table}

\begin{table*}[hb]
\centering
\small
\caption{Unified 6C scoring rubric. Clarity, Completeness, and Consistency are evaluated with LLM-based weighted heuristics, while Connectivity, Compliance, and Context Availability are computed from graph, lifecycle, and metadata signals.}
\setlength{\tabcolsep}{4pt}
\begin{tabular}{p{2.5cm}p{1cm}p{3cm}p{1cm}p{5.5cm}}
\toprule
\textbf{Dimension} & \textbf{Type} & \textbf{Heuristic / Criterion} & \textbf{Weight} & \textbf{Interpretation} \\
\midrule

\multirow{7}{*}{Clarity}
& \multirow{7}{*}{LLM}
& Structure & 0.15 & Clear subject, verb, and outcome \\
& & Quantification precision & 0.15 & Precise numerical or statistical values \\
& & Clinical specificity & 0.15 & Specific endpoints, adverse events, or targets \\
& & Context completeness & 0.15 & Population, setting, or usage context is defined \\
& & Actionability & 0.15 & Clinically or practically usable \\
& & Scientific alignment & 0.15 & Aligned with reference evidence \\
& & Completeness depth & 0.10 & Covers relevant claim elements \\

\midrule
\multirow{5}{*}{Completeness}
& \multirow{5}{*}{LLM}
& Structural integrity & 0.30 & Clear claim structure \\
& & Self-containment & 0.25 & Understandable without external context \\
& & Entity specificity & 0.15 & Key entities are explicitly defined \\
& & Outcome explicitness & 0.15 & Outcome is explicitly stated \\
& & Legal assertion clarity & 0.15 & Definitive, non-ambiguous assertion \\

\midrule
\multirow{4}{*}{Consistency}
& \multirow{4}{*}{LLM}
& Semantic alignment & 0.35 & Meaning is aligned across claim, MTV, and material \\
& & Factual consistency & 0.25 & Facts match exactly \\
& & Contextual alignment & 0.20 & Same population, indication, and scenario \\
& & Omission distortion & 0.20 & No critical information is removed \\

\midrule
Connectivity
& Data-driven
& Evidence linkage
& 1 
& Presence of active references, archived references, and linked materials \\

Compliance
& Data-driven
& lifecycle state and connectivity
& 1
& Connectivity score $\geq 0.9$ and lifecycle state is approved \\

Context Availability
& Data-driven
& Metadata completeness
& 1
& Fraction of required metadata fields populated \\

\bottomrule
\end{tabular}

\label{tab:6c_unified_rubric}
\end{table*}

\paragraph{Clarity.}
Clarity evaluates whether a claim is clearly written and scientifically precise:
\[
\begin{aligned}
s_i^{\mathrm{clar}} &=
0.15h_1 + 0.15h_2 + 0.15h_3 + 0.15h_4 \\
&\quad + 0.15h_5 + 0.15h_6 + 0.10h_7 \\
\end{aligned}
\]
\noindent\textit{Inputs: Product, Indication, Claim Text, and Reference Text.}

\paragraph{Completeness.}
Completeness evaluates whether the claim is structurally whole, self-contained, and legally anchored:
\[
s_i^{\mathrm{comp}} =
0.30h_1 + 0.25h_2 + 0.15h_3 + 0.15h_4
+ 0.15h_5
\]

\noindent\textit{Inputs: Product, Indication, Claim Text, and Reference Text.}

\paragraph{Consistency.}
Consistency evaluates semantic alignment across the claim text, Match Text Variant, and material text:
\[
s_i^{\mathrm{cons}} =
0.35h_1 + 0.25h_2 + 0.20h_3 + 0.20h_4
\]

\noindent\textit{Inputs: Product, Indication, Claim Text, Reference Text, Matched Variation Text, and Material Text.}

Let $\mathcal{M}$ denote the set of required metadata fields:
\[
\begin{aligned}
\mathcal{M} &=
\{
\textit{Indication},
\textit{ScientificArea},
\textit{PrimaryLocation}, \\
&\quad\text{ClaimCategory}, 
\textit{REFAnchorID},,\\
&\quad
\textit{MaterialLinkedText},\\
&\quad \textit{TargetAudience},
\textit{ContentType}
\}.\end{aligned}
\]
For claim $x_i$, let $m_i$ be the number of missing fields in $\mathcal{M}$. The \textbf{Context Availability score} is:
\[
s_i^{\mathrm{ctx}} =
\max\left(0, 1 - \frac{m_i}{|\mathcal{M}|}\right)
=
\max\left(0, 1 - \frac{m_i}{8}\right).
\]

\paragraph{Dependency structure.}
The data-driven dimensions are evaluated with gating dependencies:
\[
s_i^{\mathrm{conn}} = f_{\mathrm{conn}}(r_i, a_i, m_i),
\]
where $r_i$, $a_i$, and $m_i$ indicate the availability of active references, archived references, and linked materials.

\textbf{Compliance} is conditioned on connectivity:
\[
s_i^{\mathrm{compl}} =
f_{\mathrm{compl}}(s_i^{\mathrm{conn}}, \ell_i),
\]
where $\ell_i$ denotes the lifecycle state. In our implementation, a claim receives full compliance credit only when:
\[
s_i^{\mathrm{conn}} \geq 0.9
\quad \land \quad
\ell_i \in \{\text{Approved}, \text{Approved State}\}.
\]

\textbf{Context availability} is evaluated after connectivity and compliance have been established:
\[
s_i^{\mathrm{ctx}} =
\mathbb{I}[s_i^{\mathrm{conn}} \geq 0.9]
\cdot
\mathbb{I}[s_i^{\mathrm{compl}} > 0]
\cdot
\max\left(0, 1 - \frac{m_i}{8}\right).
\]

\begin{algorithm*}[t]
\small
\caption{6C Claim Quality Scoring}
\label{alg:6c_scoring}
\begin{algorithmic}[1]
\Require Claim $x_i$, reference texts $R_i$, material texts $M_i$, metadata fields $\mathcal{M}_i$, lifecycle state $\ell_i$
\Ensure Final score $S_{6C}(x_i)$

\State Retrieve linked references, archived references, materials, MTVs, and metadata from the claim graph.
\State Compute connectivity score $s_i^{\mathrm{conn}}$ from linked references, archived references, and materials.
\State Compute compliance score $s_i^{\mathrm{compl}}$ using $s_i^{\mathrm{conn}}$ and lifecycle state $\ell_i$.

\If{$s_i^{\mathrm{conn}} \geq 0.9$ and $s_i^{\mathrm{compl}} > 0$}
    \State Compute context score $s_i^{\mathrm{ctx}} = \max(0, 1 - m_i/8)$.
\Else
    \State Set or defer $s_i^{\mathrm{ctx}}$ according to the gating policy.
\EndIf

\Statex
\State In parallel, evaluate LLM-based scores:
\State $s_i^{\mathrm{clar}} \gets \mathrm{LLM}_{clarity}(x_i, R_i)$
\State $s_i^{\mathrm{comp}} \gets \mathrm{LLM}_{completeness}(x_i, R_i)$
\State $s_i^{\mathrm{cons}} \gets \mathrm{LLM}_{consistency}(x_i, R_i, M_i)$

\State Aggregate:
\[
S_{6C}(x_i) =
\frac{1}{6}
\left(
s_i^{\mathrm{clar}} +
s_i^{\mathrm{comp}} +
s_i^{\mathrm{cons}} +
s_i^{\mathrm{conn}} +
s_i^{\mathrm{compl}} +
s_i^{\mathrm{ctx}}
\right)
\]

\State \Return $S_{6C}(x_i)$
\end{algorithmic}
\end{algorithm*}

\begin{table*}[t]
\centering
\caption{Failure mode matrix for Veeva claim records. The table summarizes observed data quality and governance pain points that motivate the Golden Data Model.}
\small

\begin{tabular}{p{3cm}p{5cm}p{5cm}}
\toprule
\textbf{6C Lens} & \textbf{Observed Pain Point} & \textbf{Example} \\
\midrule

Clarity 

& Frequent use of clinical shorthand and noun-only claims, which limits standalone interpretability. 
& \texttt{CL-US-000853}: ``T1 Gd+ lesions'' lacks a verb, benefit, or complete claim structure. \\
\midrule
Connectivity 

& Claims are often linked to materials, but parent-child claim relationships remain weak, making it difficult to verify whether semantically similar claims communicate the same message. 
& 137 unique materials were tracked to claim IDs; 236 of 480 claims were referenced in at least one material. \\
\midrule
Context 
 
& Statistical or numerical claims may appear without required qualifiers or supporting context, increasing the risk of unsupported downstream generation. 
& \texttt{CL-US-000132}: numerical claims may ``float'' without the supporting messaging required for CMLR review. \\
\midrule
Consistency 

& Overlap across formulations or variants can create identity conflicts when similar wording is reused across distinct product contexts. 
& Identical VA access claims were observed for IV and SC formulations. \\
\midrule
Completeness 
 
& Long-form claims, especially safety-related claims, are prone to truncation and loss of legally relevant details. 
& \texttt{CL-US-004992}: detailed IgG monitoring instructions are vulnerable to truncation. \\
\midrule
Compliance 
 
& Core efficacy claims generally show strong reference-to-use linkage, but compliance still depends on lifecycle status and evidence availability. 
& High reference count was observed for core efficacy statements. \\

\bottomrule
\end{tabular}

\label{tab:veeva_failure_modes}
\end{table*}

\begin{table*}[t]
\centering
\caption{Definition of the 6C evaluation framework.}
\small
\begin{tabular}{p{2.5cm} p{4cm} p{3cm} p{4cm}}
\toprule
\textbf{Dimension} & \textbf{Definition} & \textbf{Core Question} & \textbf{What It Evaluates} \\
\midrule
Clarity & Linguistic clarity and interpretability & Can I understand this? & Precision, readability \\

Completeness & Self-contained structured claim & Is everything included? & Structure, entities, outcomes \\

Consistency & Alignment across claim variants & Is it consistent across versions? & Message fidelity \\

Connectivity & Strength of supporting evidence & How strong is the evidence? & Reference linkage \\

Compliance & Regulatory validity & Is it allowed to be used? & Approval + evidence \\

Context & Availability of metadata & Is required info present? & Audience, material, metadata \\
\bottomrule
\label{fig:what}
\end{tabular}

\end{table*}

\begin{figure*}[ht]
    \centering
    \includegraphics[width=1\columnwidth]{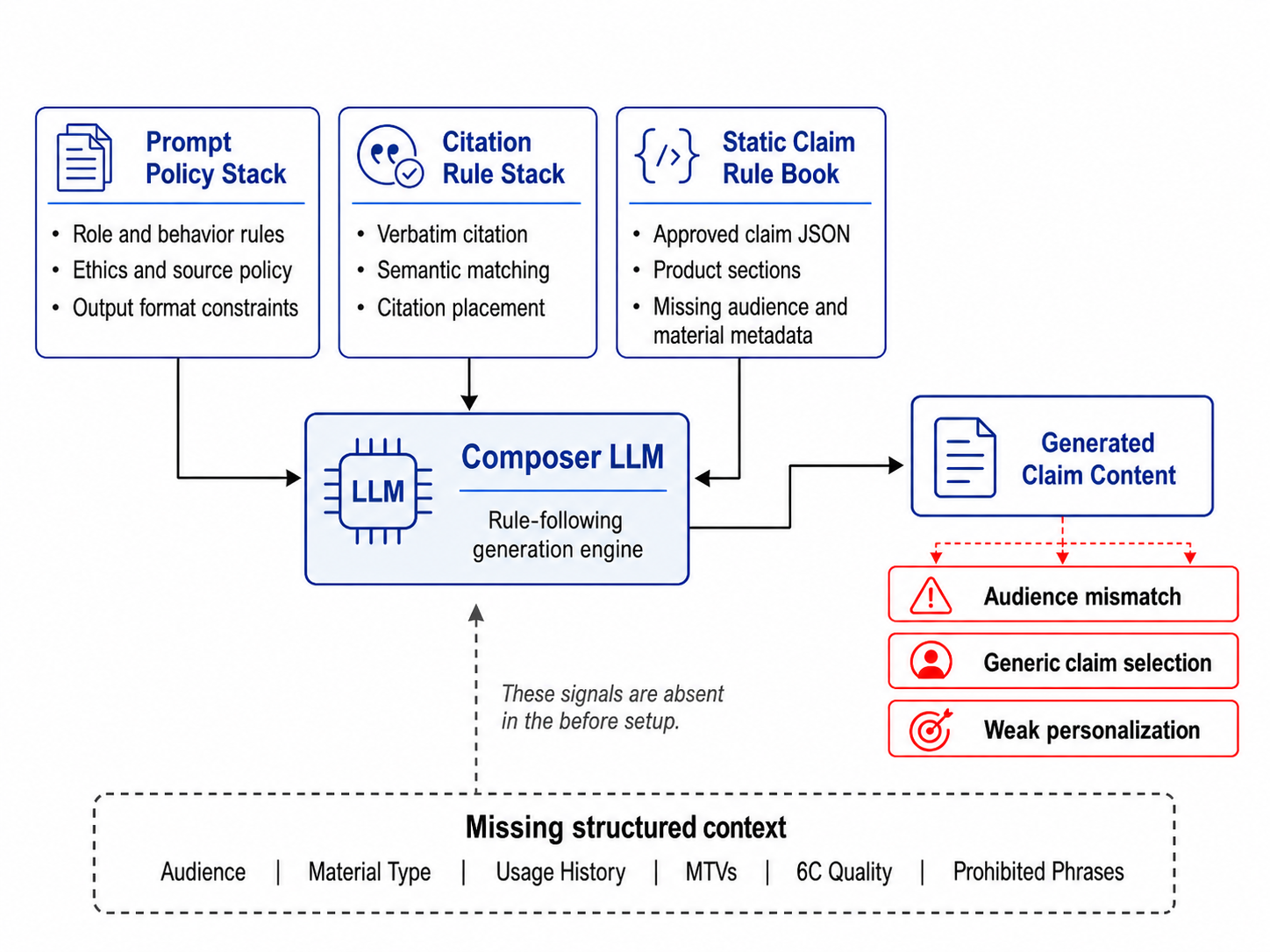}
    \caption{Rule-heavy Composer setup before the Golden Data Model. The LLM receives separate prompt policies, citation constraints, and static claim rule book. Although this design enforces basic compliance, it does not provide structured audience, material, usage, or claim-quality signals, leading to generic outputs, audience mismatch, weak personalization.}
      \label{fig:segment_improvement2}
    \vspace{-4pt}  
\end{figure*}

\begin{table*}[t]
\centering
\caption{Example claim record (CL-US-002291) represented in the Golden Data Model schema format.}
\footnotesize
\begin{tabular}{lp{10cm}}
\toprule
\textbf{Field} & \textbf{Value} \\
\midrule

Claim ID & CL-US-002291 \\

Create Date & 2023-10-17T19:48:36.000Z \\

Modified Date & 2024-08-10T14:14:09.000Z \\

Product & Ocrevus \\

Country & United States \\
\midrule
\rowcolor{rocheblue!20} Match Text (Original Claim) & OCREVUS is proven effective in reducing relapses in RMS, slowing disability progression in RMS and PPMS, and reducing brain lesions in RMS and PPMS \\
\midrule
Claim Category & Dosing Administration \\

Scientific Area & Neuroscience \\

Indication & Multiple Sclerosis \\
\midrule
\rowcolor{rocheblue!20}Lifecycle State & Approved \\
\midrule
Source Claim ID & -- \\

Reference IDs & [LT-018562, LT-018566, LT-033069, LT-033060, LT-033147, LT-033129, LT-033185, LT-033159] \\

Reference Texts & Clinical trials demonstrate efficacy in relapse reduction, disability progression delay, and MRI lesion reduction across RMS and PPMS populations. \\

Content Type & Social Media Post \\
\midrule
\rowcolor{rocheblue!20}Target Audience & Patient \\
\midrule
AI Guardrails Hallucination Prevention & 
\texttt{[evidence\_alignment: keep wording aligned to reference text; audience\_alignment: patient-friendly language; claim\_category\_alignment: do not change dose, timing, route, or preparation details]} \\
\midrule
Claim Specific Business Rule & 
\texttt{\{tone\_signals: [patient-friendly, consumer-facing, instructional, outcome-oriented], primary\_message\_type: Dosing Administration\}} \\
\midrule
Prohibited Words Negative Constraints & 
\texttt{\{important\_phrases: [OCREVUS, RMS, PPMS, proven effective], avoid: [absolute benefit claims, guaranteed outcomes, cure or reverse disease, modifying dosing details], status: human\_review\}} \\
\midrule
6C Score & 1.00 \\

\bottomrule
\end{tabular}
\label{tab:example_claim_full}
\end{table*}

\begin{figure*}[hb]
    \centering
    \begin{minipage}{1\textwidth}
        \centering
        \includegraphics[width=1\textwidth]{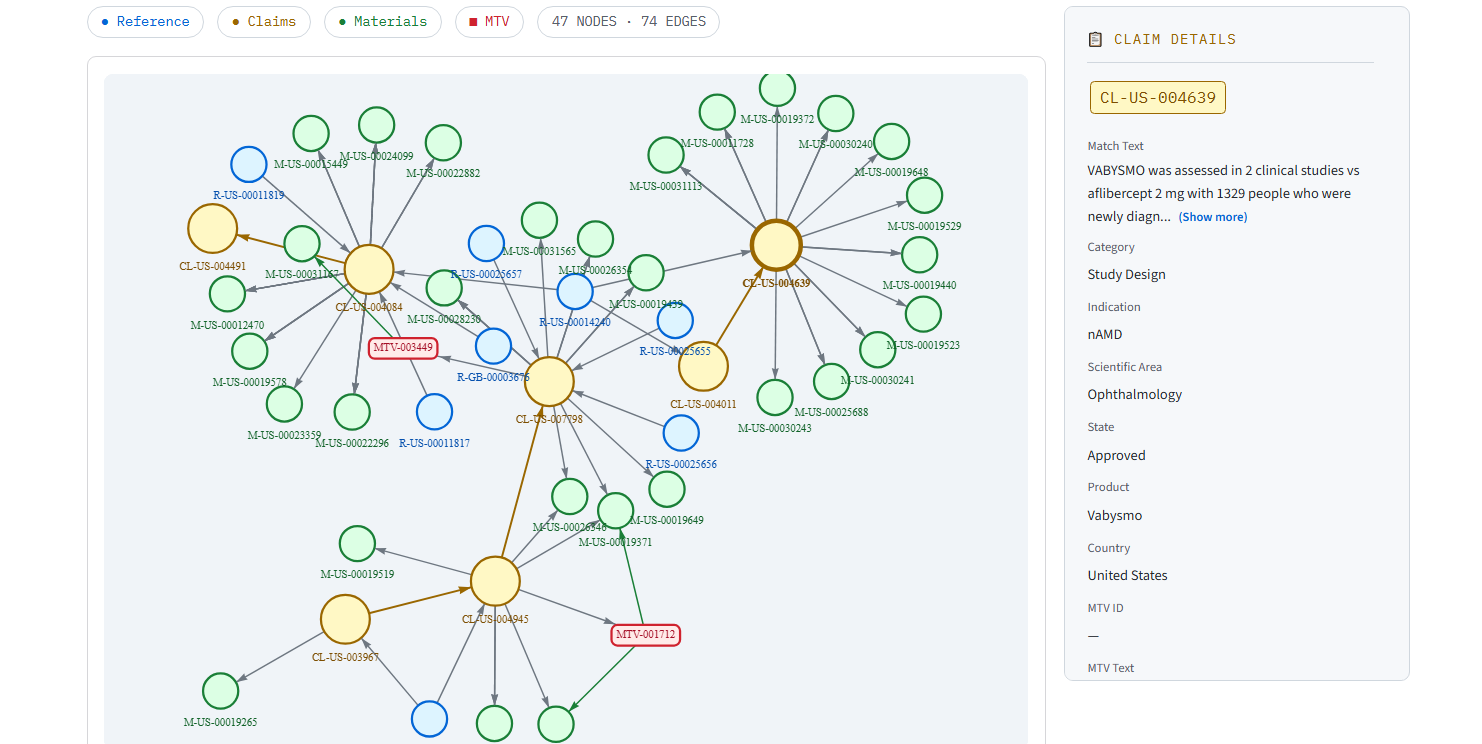} 
    \end{minipage}%
    
    \caption{Comprehensive Graph View of Claims.}
        \label{fig: architecture of graph}
\end{figure*}

\begin{figure*}[t]
\centering
\begin{minipage}{\textwidth}
\begin{lstlisting}[style=jsonstyle]
{
  "documentInfo": {
    "title": "<product> Approved Claims Analysis",
    "content_about": "<product-level claim summary> e.g., This document contains a compilation of approved claims for the drug Ocrevus, covering various categories such as Dosing & Administration, Efficacy, Safety, Mechanism of Action, Disease Education, Study Design, Utilization, and others. It details rules, guidelines, and findings from clinical trials and real-world data",
    "other_comments": "<coverage notes and data quality limitations> e.g., The document aggregates 847 claims across 7 primary categories, with additional claims grouped into other relevant sections. Based on the provided context, a significant portion of the source data for these claims reportedly lacks complete information, such as claim category or assigned agency. Claims without a clear, explicit category in the source text have been contextually assigned or placed in an 'Others' section."
  },
  "sections": [
    {
      "title": "<section title, e.g., 1. Dosing & Administration>",
      "content_about": "<section-level summary>, e.g., 1.1 This section details the rules and guidelines for the preparation, administration, and management of Ocrevus infusions. It includes information on patient counseling, premedication, dosing schedules, infusion rates, storage, and handling of infusion-related reactions (IRs)",
      "what_to_do": [
        "<approved rule or claim 1>, e.g., 1.1 Contraception: Advise female patients of childbearing potential to use effective contraception while receiving OCREVUS and for 6 months after the last infusion of OCREVUS",
        "<approved rule or claim 2>",
        "..."
      ],
      "what_to_prohibit": [
        "<prohibited wording or constraint 1>",
        "..."
      ]
    }
  ]
}
\end{lstlisting}
\end{minipage}

\caption{\textbf{BEFORE}: Legacy product-level rule JSON used in the rule-based Composer setup. The structure stores product summaries and section-level rule lists, but lacks explicit audience, material type, usage history, MTV, and claim-quality signals required for personalized claim generation.}
\label{fig:legacy_rule_json}

\end{figure*}

\begin{table*}[ht]
\centering
\caption{Foundation Prompt System Instruction.}
\footnotesize
\begin{tabular}{p{0.20\linewidth} p{0.75\linewidth}}
\toprule
\rowcolor{rocheblue!20}\textbf{System Prompt} & \textbf{Content} \\
\midrule

Role and Context
&
Process claim-level data into a structured data model. Extract information from each claim match text to populate \texttt{layer\_ai\_guardrails\_hallucination\_prevention}. Derive \texttt{claim\_specific\_business\_rule} by combining the reference text and match claim text. Populate \texttt{marketing\_vector\_store\_content} with tone signals such as patient-friendly, benefit-led, or HCP-focused. The \texttt{primary\_message\_type}, such as Efficacy, should be inherited from the claim classification. Populate \texttt{prohibited\_words\_negative\_constraints} by extracting important phrases and phrases to avoid or not overstate from the original claim text. \\

\midrule

Output Format
&
Output should be an Excel file. Keep the output structured at the claim-row level, not one row per unique claim ID. Include the following fields: \texttt{layer\_ai\_guardrails\_hallucination\_prevention}, \texttt{claim\_specific\_business\_rule}, \texttt{marketing\_vector\_store\_content}, and \texttt{prohibited\_words\_negative\_constraints}. \\

\midrule

Tool and Processing Policy
&
Do not rewrite the claims themselves. Match each row by its own claim ID and target audience combination. Do not collapse multiple rows into one row per claim ID, since the true unique unit is the combination of claim ID and target audience. \\

\midrule

AI Guardrails Hallucination Prevention
&
\texttt{\{``rule\_type'': ``evidence\_alignment'', ``derived\_from'': [``Reference Texts'', ``Match Text''], ``rule'': ``Keep wording aligned to the match text and supported by the available reference text.''\}} \\

\midrule

Alignment 
&
\texttt{\{``rule\_type'': ``audience\_alignment'', ``target\_audience'': ``Patient/Consumer-facing'', ``rule'': ``Prefer plain-language wording and avoid adding HCP-only terminology unless already present in the source text.''\}}
 \\
&
\texttt{\{``rule\_type'': ``material\_alignment'', ``material\_type'': ``Post Service / Mail'', ``rule'': ``Keep the wording appropriate for the material type without changing claim meaning.''\}}
 \\
&
\texttt{\{``rule\_type'': ``claim\_category\_alignment'', ``claim\_category'': ``Efficacy'', ``rule'': ``Retain benefit wording without upgrading to guaranteed, universal, or curative outcomes.''\}} \\

\midrule

Marketing Vector Store Content Analysis
&
\texttt{\{``tone\_signals'': [``patient-friendly'', ``consumer-facing'', ``HCP-focused'', ``clinical'', ``web-ready'', ``email-ready'', ``benefit-led'', ``outcome-oriented'', ``explanatory''], ``primary\_message\_type'': ``Efficacy''\}} \\

\midrule

Prohibited Words and Negative Constraints
&
\texttt{\{``source'': ``derived\_from\_match\_text'', ``exact\_claim\_anchor'': ``VABYSMO can help quickly dry vision-damaging fluid in your eye'', ``important\_phrases\_from\_claim'': [``VABYSMO'', ``eye'', ``quickly dry vision-damaging fluid''], ``phrases\_to\_avoid\_or\_not\_overstate'': [``remove qualifier such as can'', ``remove qualifier such as help'', ``overstating efficacy beyond the source wording'', ``medical advice replacing a healthcare professional'', ``overly promotional language not supported by the source text''], ``status'': ``derived\_needs\_human\_review''\}} \\

\bottomrule
\end{tabular}

\label{tab:foundation_qc_rules}
\end{table*}

\clearpage
\onecolumn
\raggedbottom

\subsection{Ablation Study: Post-Deployment Composer Usage Patterns} \label{sec:A3} 

% Figure on the LEFT; text wraps on the RIGHT
\begin{wrapfigure}[38]{l}{0.47\textwidth}
 
  \centering

  \captionsetup{
    font=small,
    labelfont=bf,
    justification=justified,
    singlelinecheck=false,
    aboveskip=3pt,
    belowskip=0pt
  }

  \includegraphics[width=\linewidth]
    {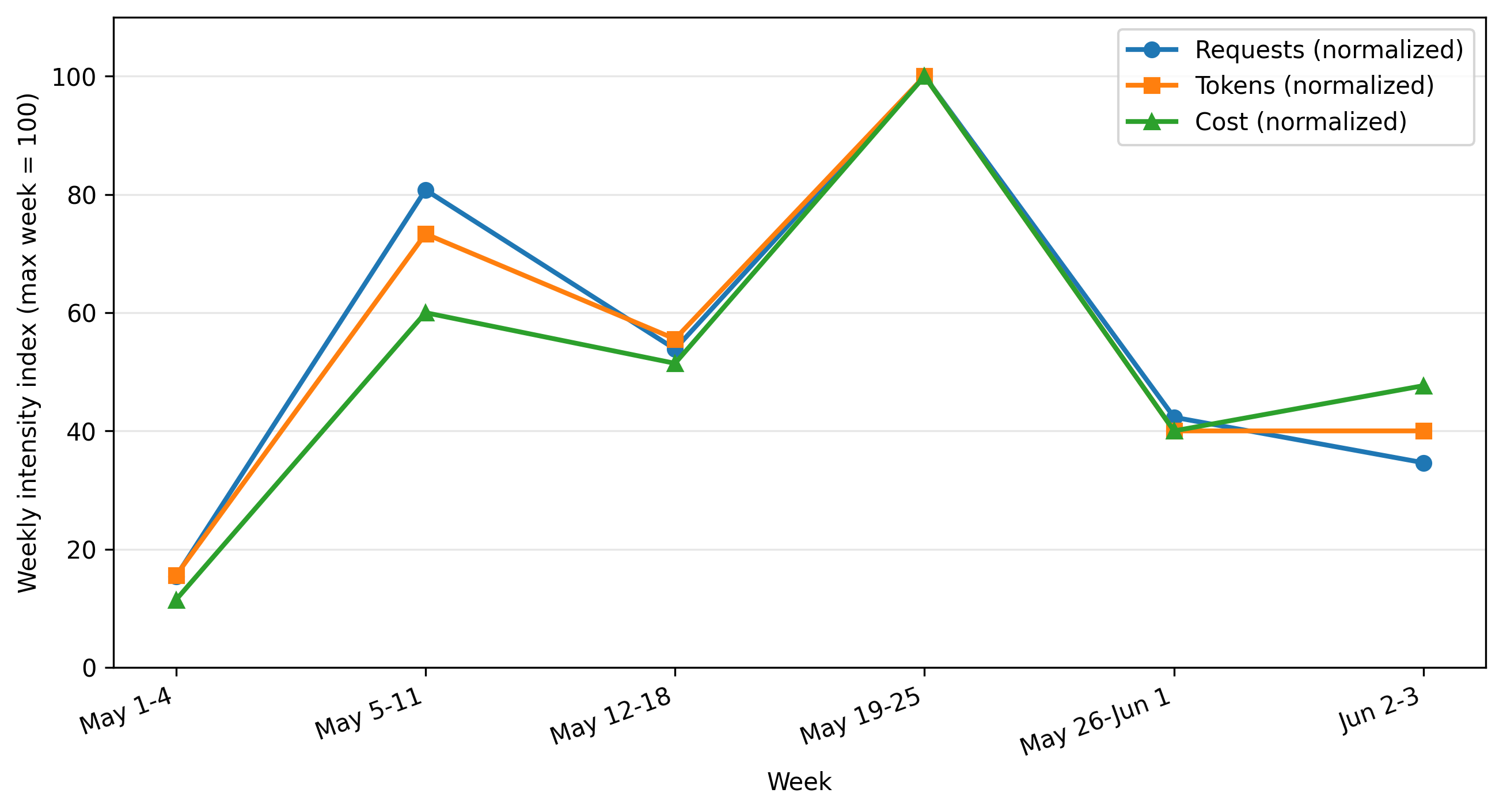}

  \vspace{-3pt}
  {\centering\footnotesize
  \textbf{(a)} Weekly intensity index.\par}

  \vspace{3pt}

  \includegraphics[width=\linewidth]
    {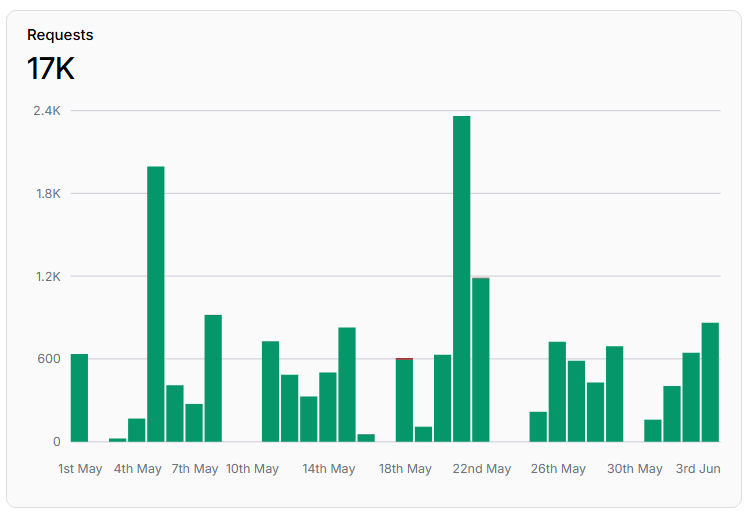}

  \vspace{-3pt}
  {\centering\footnotesize
  \textbf{(b)} Requests.\par}

  \vspace{3pt}

  \includegraphics[width=\linewidth]
    {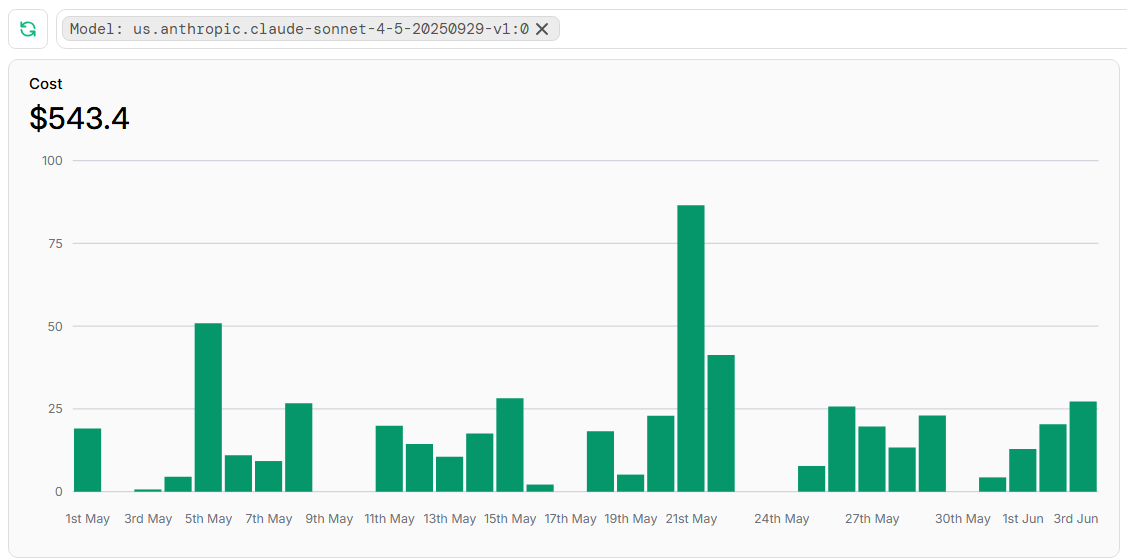}

  \vspace{-3pt}
  {\centering\footnotesize
  \textbf{(c)} API cost.\par}

  \vspace{3pt}

  \includegraphics[width=\linewidth]
    {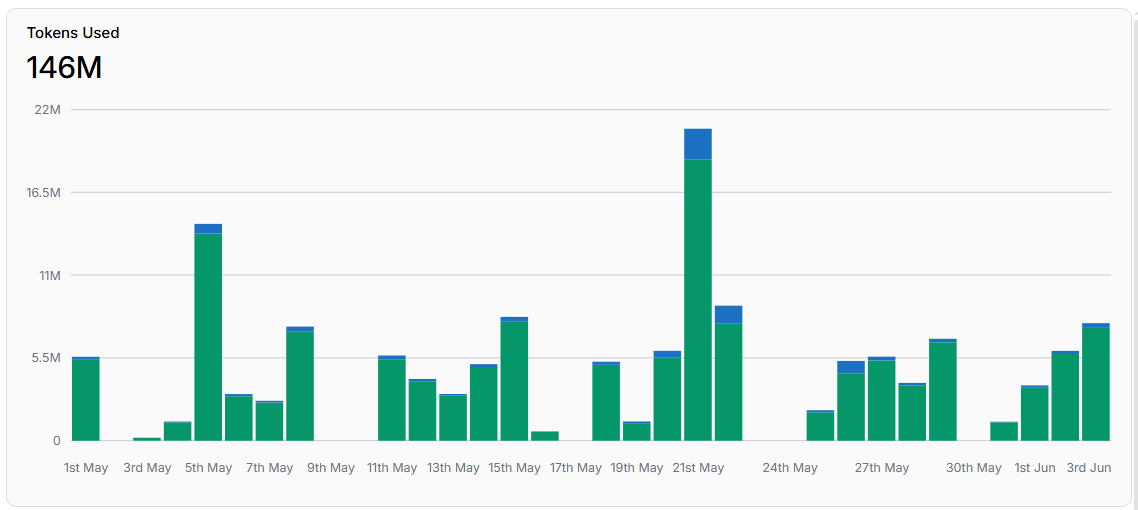}

  \vspace{-3pt}
  {\centering\footnotesize
  \textbf{(d)} Token usage.\par}

  \caption{\textbf{Post-deployment Composer usage patterns.}
  The analyzed logs contain 16,963 requests, 146 million tokens,
  and an estimated monthly API cost of \$543.39. The temporal
  distribution suggests human usage.}
  \label{fig:finding}

\end{wrapfigure}

\textbf{Problem:}Can Composer generate emails and banners that better match the target audience, user intent, indication, and content format while reducing production API costs? 

\textbf{What we did:} We compared the original rule-based pipeline with a new context-based pipeline powered by the Golden Data Model. The original pipeline primarily relied on prompt instructions and approved-claim rules. In contrast, the Golden Data Model provides structured claim-level context, including references, audience labels, MTV links, 6C scores, and LLM-derived tags for hallucination prevention and prohibited wording. 

\textbf{Outcome:} The logs contain \textbf{16,963 requests}, \textbf{146 million tokens},and an estimated \textbf{monthly API cost of \$543.39}. We evaluated model-side measures, including input and output tokens, latency, and success rate, and user-side measures, including daily activity, request volume, and likely regeneration behavior. The results indicate systematic human usage rather than random API traffic. \textbf{Wednesdays generally show lower activity, whereas Mondays, Tuesdays, and Fridays show higher activity}, which helps identify safer windows for system updates and support planning. Our context-based pipeline also required fewer repeated requests and increased the target-audience match rate by \textbf{67\%}.

\end{document}